\documentclass[11pt,a4paper]{article}
\usepackage[hyperref]{acl2020}
\usepackage{times}
\usepackage{latexsym}

\usepackage{microtype}

\usepackage{amsmath,amsthm,amssymb,amsfonts}
\usepackage{booktabs}
\usepackage{graphicx} 
\usepackage{bbm}
\usepackage{comment}
\usepackage{enumitem}
\usepackage{xcolor}

\usepackage{multirow}

\aclfinalcopy

\title{Consistency and Coherency Enhanced Story Generation}

\author{Wei Wang$^{1,}$\footnotemark[1]\thanks{~~Work was done during internship at Tencent AI Lab.}~, Piji Li$^2$, Hai-Tao Zheng$^1$ \\
$^1$Shenzhen International Graduate School, Tsinghua University \\ 
$^2$Tencent AI Lab \\
 {\tt w-w16@mails.tsinghua.edu.cn} \\
  {\tt pijili@tencent.com} \\
   {\tt zheng.haitao@sz.tsinghua.edu.cn}
}

\begin{document}

\maketitle
\begin{abstract}
	Story generation is a challenging task, which demands to maintain consistency of the plots and characters throughout the story. Previous works have shown that GPT2, a large-scale language model, has achieved good performance on story generation. However, we observe that several serious issues still exist in the stories generated by GPT2 which can be categorized into two folds: consistency and coherency. In terms of consistency, on one hand, GPT2 cannot guarantee the consistency of the plots explicitly. On the other hand, the generated stories usually contain coreference errors. In terms of coherency, GPT2 does not take account of the discourse relations between sentences of stories directly. To enhance the consistency and coherency of the generated stories,  we propose a two-stage generation framework, where the first stage is to organize the story outline which depicts the story plots and events, and the second stage is to expand the outline into a complete story. Therefore the plots consistency can be controlled and guaranteed explicitly. In addition, coreference supervision signals are incorporated to reduce coreference errors and improve the coreference consistency. Moreover, we design an auxiliary task of discourse relation modeling to improve the coherency of the generated stories.  Experimental results on a story dataset show that our model outperforms the baseline approaches in terms of both automatic metrics and human evaluation.
	
% \keywords{Natural language generation  \and Story generation \and Language model.}
\end{abstract}

\section{Introduction}

A story is anything which is narrated in the form of a causally/logically linked set of event plots involving some shared characters \cite{mostafazadeh-etal-2016-corpus}. Story generation aims at automatically generating stories with such attributes.
% , i.e., plots and characters consistency as well as the content coherency. 
Different from other text generation tasks, story generation is more challenging because it is restricted to several constraints: (1) it must maintain \textbf{consistent plots} to form a reasonable story; (2) it must guarantee the \textbf{consistency} of the characters throughout the story; (3) the \textbf{coherency} of the text units such as the clauses or sentences should be concerned. In order to improve the quality of plot consistency, some previous works focus on the perspective of plot planning and then merge the text units according to the order of plots \cite{lebowitz1987planning,perez2001mexica,porteous2009controlling,riedl2010narrative,li2013story}. We observe that those approaches rely heavily on human annotations and are restricted to the abstract story representation level without surface realization in natural language, such as producing event verb sequence and sentence segments. Therefore, these methods need to work with sentence templates or rules to generate stories.

\begin{table}[!t]
	\centering
% 	\resizebox{1\columnwidth}{!}{
\begin{tabular}{p{0.98\columnwidth}} 
	\toprule
	\small \textbf{Prompt}:  The zombie apocalypse is upon us but all humans are immune to the virus . All other mammals are susceptible \\
	\midrule
	\small \textbf{Story}: As I sat there I tried to remember how many days it had been since the last one died . $[$\textcolor{blue}{My wife and daughter were dead , I knew they would n't be alive any time soon .}$]_{1a}$ The food ran out quickly , \textcolor{red}{they} had to be taken away to stay in the house , but that was no reason to be angry if there were a few people left living there . After a month of food being gone a little girl sat next to me . She was the only other survivor who was n't bitten , as it was hard for us to eat \textcolor{red}{them} , we were all sickly . That would be another month to go and if the other animals were still alive in the food , they would n't be able to eat her . `` Hey little one . '' I looked up and saw her . $[$\textcolor{blue}{I had a big smile on my face at this point .}$]_{2a}$ $[$\textcolor{blue}{My wife was the only one who noticed , but her smile was all that was visible of me , and not of her .}$]_{1b}$ $[$\textcolor{blue}{I looked at her , and smiled sadly .}$]_{2b}$ She looked sad , but did n't care . She never cared for me .
	\\
	\bottomrule
	\caption{A story generated by GPT2.}
	\label{case:storyexample}
\end{tabular}
% }
 \vspace{-11mm}
\end{table}

In the past few years, several end-to-end approaches based on Sequence-to-Sequence (Seq2Seq) models \cite{sutskever2014sequence,bahdanau2014neural} are proposed, which can generate a story at a stroke in a left-to-right manner \cite{jain2017story,clark2018neural,fan2018hierarchical}. These methods are data-driven and can directly generate stories in natural language form instead of other abstract representation. However, these methods struggle to capture the high-level interactions between the plot points and maintain consistent plots throughout the story. Thus, several two-stage models for story generation have recently been proposed \cite{martin2018event,xu2018skeleton,yao2019plan,fan-etal-2019-strategies,chen2019learning}. These models usually decompose story generation into two stages: generating middle form first and then generating the final story. Different middle forms are applied in these methods, such as keywords, sentences and event tuples.

Recently, the OpenAI GPT2/3 language model \cite{radford2019language,brown2020language} achieves strong performance on several language generation tasks. \cite{see-etal-2019-massively} and \cite{guan2020knowledge} verify the performance of GPT2 on story generation and GTP2 outperforms both end-to-end methods and two-stage methods.  
However, after analyzing the generated stories carefully, we observe that there are still some serious issues in the generated stories by GPT2. 
Take a story generated by GPT2 as shown in Figure \ref{case:storyexample} for example. The story is about survivors in the end of the world. First, plots consistency cannot be guaranteed among multiple sentences of a story, such as blue sentences in Figure \ref{case:storyexample}. The sentence $1a$ describes ``My wife and daughter were dead ''. But the sentence $1b$ talks about ``My wife'' again. It is contradictory. There is the same problem in the sentence $2a$ and $2b$. Second, there are still coreference errors in generated stories, such as red text in Figure \ref{case:storyexample}. It is not clear who \textit{they} and \textit{them} refer to.
Moreover, Top-k sampling~\cite{radford2019language,see-etal-2019-massively,brown2020language} is usually utilized as the decoding strategy in long text generation. The random operation in sampling will disturbance the generation procedure by producing improper tokens which will decrease the quality. This phenomenon is more pronounced at the border of sentences, therefore we can sometimes observe the bad performance in discourse coherency.     

To solve the aforementioned problems, we propose a two-stage generation model based on Transformer-based auto-regressive language models to improve consistency and coherency of stories. Specifically,  the first stage is to organize the story outline which depicts the story plots and events, and the second stage is to expand the outline into a complete story. Therefore the plots consistency can be controlled and guaranteed explicitly. In addition, coreference supervision signals are incorporated to reduce coreference errors and improve the coreference consistency.
Moreover, we design an auxiliary task of discourse relation modeling to enhance the discourse coherency of the generated stories. 
Both the backbone models in the two states are designed based on Transformer-based language models. Thus, on one hand, the framework can still inherit the superior performance of GPT2, on the other hand, it can guarantee the plot consistency, coreference consistency, as well as discourse coherency.

The main contributions of this paper are summarized as follows:
\begin{itemize}[topsep=0pt]
\setlength\itemsep{-0.4em}
	\item We propose to improve the plot and coreference consistency as well as the discourse coherency for the task of story generation.
	\item A two-stage  framework based on Transformer-based language models is designed to control the plots and improve consistency of generated stories. 
	\item A coreference constraint is applied to improve the coreference consistency of generated stories.
	\item We design a discourse relation modeling component as an auxiliary task during training to enhance the performance of discourse coherency.
	\item Experiments on a story dataset from Reddit demonstrate that our model outperforms the baseline methods in terms of both automatic metrics and human evaluation.
\end{itemize}

%The remainder of this paper is organized as follows. We review related
%works in Section 2. Section 3 introduces the detail of our proposed model. Section 4 describes dataset, experimental setups and gives the result analysis. Finally, we give a summary of this paper in Section 5.

\section{Methodology}

\subsection{Overview}

\begin{figure*}[!ht]
	\centering
	\includegraphics[width=15cm]{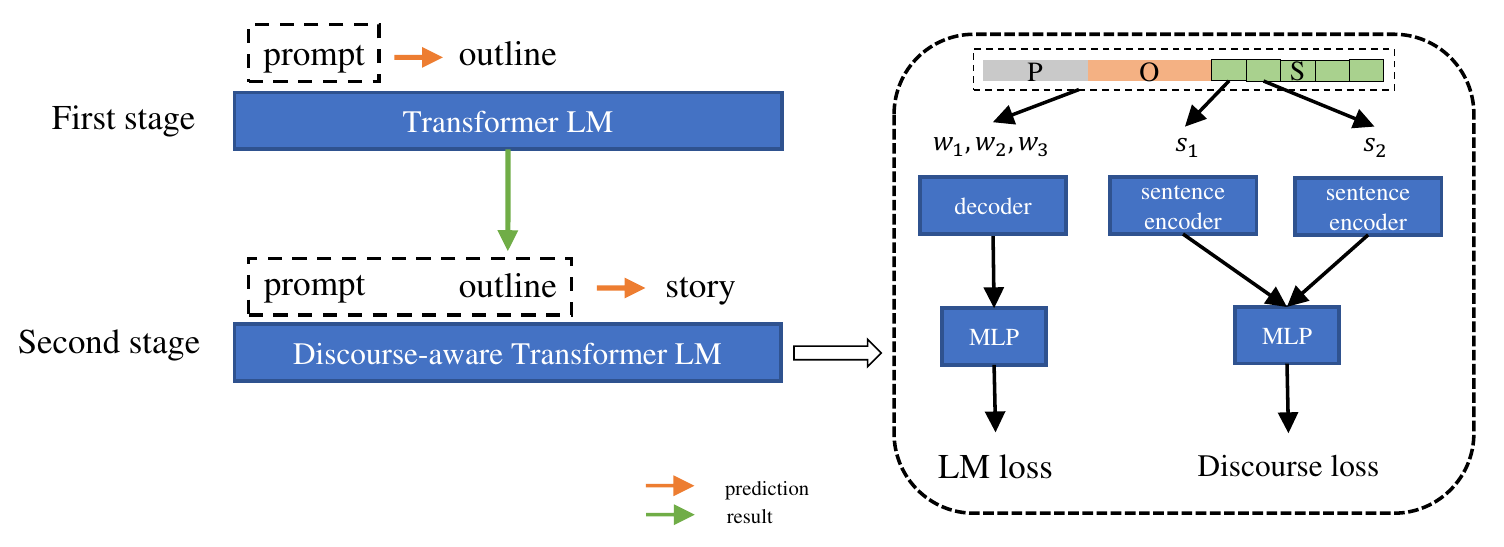}
	\caption{The framework of our model for story generation.}
	\label{fig:allmethod}
% 	\vspace{-4mm}
\end{figure*}

To begin with, we state the problem of story generation as follows: given a prompt context $\mathbf{X} = \{x_1,...,x_i...,x_k\}$ where $x_i$ denotes each word in the prompt, the model needs to generate a story $\mathbf{Y} = \{y_1,...,y_i,...,y_n\}$ following the prompt $\mathbf{X}$ by maximizing the conditional probability $p(\mathbf{Y}|\mathbf{X})$.

As show in Figure \ref{fig:allmethod}, to enhance the consistency and coherency of generated stories, we propose a two-stage framework for story generation. The first stage is story outline generation which can generate the plot outline based on the given prompt. Then in the second stage, the whole story is completed by embellishing the outline generated in the first stage.
Transformer-based language models are introduced as the backbone models for those two stages respectively.
Moreover, a component of discourse relation classification is incorporated to the language model as an auxiliary task to further improve the coherency of the generated stories.
To further improve the consistency, we design a coreference supervision component to encourage the language model to attend on correct entities when generating pronouns by maximizing the attention weights of the corresponding entities. 

\subsection{Transformer-based Language Model}
Inspired by the popular pre-trained language models for text generation such as GPT2 \cite{radford2019language}, XLNET \cite{yang2019xlnet} and GPT3 \cite{brown2020language}, we also employ the Transformer-based auto-regressive language models as our backbone frameworks.

Transformer-based language models only contain a decoder. The decoder consists of $N$ identical self attention blocks and each block contains two sub-layers: a self multi-head attention layer and a feed-forward layer. A add \& norm layer is employed around each of two sub-layers. Formally, given the input $\mathbf{H}^{n-1}$, the output $\mathbf{H}^{n}$ of each decoder block is computed as follows:
\begin{align}
	\mathbf{C}^{n} & = \operatorname{LN}\left(\operatorname{SELF-ATT}\left(\mathbf{H}^{n-1}\right)+\mathbf{H}^{n-1}\right) \\
	\mathbf{H}^{n} &=\operatorname{LN}\left(\operatorname{FFN}\left(\mathbf{C}^{n}\right)+\mathbf{C}^{n}\right)
\end{align}
where  $\operatorname{SELF-ATT}(\cdot)$,  $\operatorname{LN}(\cdot)$,  and  $\operatorname{FFN}(\cdot)$ are  respectively  self-attention  mechanism,  layer normalization, and feed-forward network with ReLU activation in between.  $\operatorname{SELF-ATT}(\cdot)$ computes attention over the input $\mathbf{H}^{n-1}$ as follows:

\begin{equation}
    \operatorname{SELF-ATT}\left(\mathbf{H}^{n-1}\right)=\operatorname{softmax}\left(\frac{\mathbf{Q} \mathbf{K}^{\top}}{\sqrt{d_{k}}}\right) \mathbf{V}
\end{equation}
where $\{\mathbf{Q,K,V}\}$ are query, key and value vectors that are transformed from the input $\mathbf{H}^{n-1}$. $\sqrt{k}$ is the scaling factor where the $d_k$ is the dimension size of the query and key vectors. Given the word embeddings $\mathbf{E}=\{e_1,e_2,...,e_m\}$ and corresponding positional embeddings $\mathbf{P}=\{p_1,p_2,...,p_m\}$, the first block input $\mathbf{H}^0=\mathbf{E}+ \mathbf{P}$. 

Finally, a linear function with $\operatorname{softmax}$ activation is used to compute the probability of next word $x_t$ via:
\begin{equation}\label{eq:outprob}
	p \left( x_t | x_{\leq{t-1}}  \right) = \operatorname { softmax } \left( g \left( h _ { t } \right) \right)
\end{equation}
We calculate negative log-likelihood loss for model training:
\begin{equation}
	\mathcal{L}_{\mathrm{lm}}=- \frac{1}{T}\sum_t \log p\left(x_t |x_{\leq t-1} \right)
\end{equation}

\begin{table*}[!ht]
	
	\centering
	\resizebox{2.05\columnwidth}{!}{
	\begin{tabular}{lcl}
		\toprule
		S1                             & marker  & S2                                                \\
		\midrule
		Her eyes flew up to his face.  & and     & Suddenly she realized why he looked so different. \\
		The concept is simple.         & but     & The execution will be incredibly dangerous.       \\
		You used to feel pride.        & because & You defended innocent people.                     \\
		Belter was still hard at work. & when    & Drade and barney strolled in.                     \\
		I' ll tell you about it. & if    & You give me your number.                     \\
		We plugged bulky headsets into the dashboard. & so    & We could hear each other when we spoke into the microphones.                     \\
		It was mere minutes or hours. & before    & He finally fell into unconsciousness.                     \\
		And then the cloudy darkness lifted. & though    & The lifeboat did not slow down.                     \\
		\bottomrule
	\end{tabular}
	}
	\caption{Example pairs from  Books 8 dataset.}
% 	\vspace{-8mm}
	\label{dismaker}
\end{table*}

\subsection{Two-stage Generation}

\noindent\textbf{Outline Preparation}

% \begin{comment}
% In order to improve the coherence of the story, we propose a two-stage method as shown in Fig \ref{fig:allmethod}, which first generate a outline and then generate a complete story according to the outline. Through this two-stage approach, stories are generated under guided of outline and have better coherence. In addition, we can control plots of story by outline. What's more, for each stage we apply a Transformer-based language model decoder. We cast each stage as a language model task other than a Seq2Seq task. In this way, we can take advantage of the powerful generation ability of Transformer-based language model at each stage.
% \end{comment}

In order to regard the outline generation task as a supervised learning problem, we must construct a high-quality training dataset including sufficient prompt-outline pairs. 
As pre-mentioned, outline implies the story plots, therefore the quality of outline can affect the performance of story generation directly.
If the outline contains too much information, the story generator will directly learn to copy from the outline and restrain the imagination and creativity.
On the contrary, if the outline ignores the key-point information, the informativeness and consistency of stories will be decreased.

In this work, we investigate two forms of outline: keyword and abstract. These two forms retain the important information of the story and ignore some details and commonly used in two-stage based methods \cite{yao2019plan,fan-etal-2019-strategies,chen2019learning}. Our motivation is to use two-stage generation to improve performance of GPT2 , so we do not design a new middle form. Specifically, we use the RAKE algorithm \cite{rose2010automatic} \footnote{https://pypi.org/project/rake-nltk/} to extract keywords of story. According to \cite{yao2019plan} and the average lengths of stories in our corpus, we extract 10 keywords for each story. We use a variation of the TextRank algorithm \cite{barrios2016variations} \footnote{https://radimrehurek.com/gensim/} to extract abstract of story. In order to retain important information and ignore some detail information, we keep 30\% sentences of each story as abstract. Thus, we can get (prompt, outline, story) pairs automatically to train the two-stage model.

% \vspace{6pt}
\noindent\textbf{Prompt to Outline Generation}

% \piji{draw a fig, refer LI Xin's AAAI2020 paper Figure 1: https://arxiv.org/abs/1911.11489, to show the input, output, as well as the loss tokens, add more details, formulas, etc.}

A Transformer-based language model based decoder is used to generate outlines. Specifically, we concatenate prompt $\mathbf{X}$ and outline $\mathbf{Z}$ with  \verb|<SEP>| token to get a sequence $\mathbf{X'}$. For training, we compute cross entropy of all tokens in $\mathbf{X'}$ as normal language model. When testing, given the prompt tokens as context, the decoder generates outline tokens.

% \vspace{6pt}
\noindent\textbf{Prompt and Outline to Story Generation}
% \piji{draw a fig, refer LI Xin's AAAI2020 paper Figure 1: https://arxiv.org/abs/1911.11489, to show the input, output, as well as the loss tokens, add more details, formulas, etc.}

Another decoder with the same architecture is used to generate stories. We concatenate prompt $\mathbf{X}$, outline $\mathbf{Z}$ and story $\mathbf{Y}$ with \verb|<S>| and \verb|<SEP>| token to get a sequence $\mathbf{X''}$. For training, we compute cross entropy of prompt and story tokens in $\mathbf{X''}$. Note that we don't calculate the loss of the outline tokens. Because,  the tokens come from the story and we avoid computing loss of these tokens twice. When testing, given the prompt and the outline tokens as context, the decoder generates story tokens. Next, two components are incorporated in this stage to enhance discourse coherency and coreference consistency.

% \subsection{Coherence Enhancing via Discourse Modeling}
\subsection{Discourse Coherency Enhancement}\label{sec:discourse}
In order to improve discourse representation of Transformer-based language model, we design a discourse relation classification task as an auxiliary task. Discourse relations describe how two segments (e.g. clauses, sentences, and larger multi-clause groupings) of discourse are logically connected. These relations can be used to describe the high-level organization of text. Thus, discourse relation is an important aspect of story coherence. In this work, we only consider shallow discourse relations between adjacent sentences as many research on discourse relation classification do \cite{chen-etal-2016-implicit,lan-etal-2017-multi,bai-zhao-2018-deep}.

% \vspace{-4mm}
\noindent \textbf{Discourse Information Preparation}

In order to get discourse label of adjacent sentences in stories, we need to train a golden discourse relation classification model. However, there is limited annotation corpus of implicit discourse relations and explicit discourse relations. For example, the commonly used dataset Penn Discourse Treebank 2.0 \cite{prasad-etal-2008-penn} contains about 10k pairs. Following \cite{nie-etal-2019-dissent}, we use discourse markers as replace of discourse relations. Because we are able to automatically curate a sizable training set of sentence pairs with discourse markers. We use discourse marker dataset Book 8 from \cite{nie-etal-2019-dissent}, which contains 3.6M sentence pairs and each pair is labeled with one connective of 8 connectives as discourse label. Several sentence pairs and corresponding discourse markers are shown in Table \ref{dismaker}.

We fine tuning BERT \cite{devlin2018bert} \footnote{https://github.com/huggingface/transformers} in this dataset to get a golden discourse marker prediction model. Then we use this model to tag discourse relation label of sentence pairs in our story corpus. Considering that this automatic tagging may produce large errors, we only keep labels with high classification probability, and labels with lower probability are replaced with the ninth label, \textit{unknown}. The sentence pairs with labels belonging to 8 connectives are used to train our discourse relation classification component.

\noindent \textbf{Discourse-aware Story Generation}

The discourse relation classification component contains a sentence encoder and a two-layers MLP. The encoder is used to extract sentence semantic feature and the MLP is used to convert feature into classification probability.  The sentence encoder shares parameters with the story decoder exclude the output layer. For a story $\mathbf{Y}$ contains several sentence $\{\mathbf{S_1},\mathbf{S_i},\mathbf{S_p}\}$ and each sentence contains several words $\mathbf{S_i}=\{y_{i1},y_{ij},y_{iq}\}$, we get output $h^w_{ij}$ of encoder as word representation and use max pooling operation on words of this sentence to get sentence representation $h^s_i$:
\begin{align}
	\mathbf{H}^s_i & = \operatorname{encoder}(\mathbf{S_i})  \\
    	h^s_i &= \operatorname{max}(\mathbf{H}^s_i)
\end{align}
Then the MLP is used to classify adjacent sentences as follows:
\begin{eqnarray}
	f = &\tanh(\mathbf{W_f}[h^s_i,h^s_{i+1}]+b_f) \\
	p(dis|\mathbf{S_i},\mathbf{S_j}) = &  \operatorname{softmax}(\mathbf{W_o}f +b_o)
\end{eqnarray}
The loss function $\mathcal{L}_{\mathrm{dis}}$ of this component is the cross-entropy of discourse label. Then a joint loss function is applied to train the second stage model:
\begin{equation}
	\mathcal{L}=\mathcal{L}_{\mathrm{lm}}+ \lambda_1\mathcal{L}_{\mathrm{dis}}
\end{equation}
where $\lambda_1$ is a hyperparameter to balance two tasks.

\subsection{Coreference Consistency Enhancement}

Although Transformer-based language model has the ability of long-distance dependence, there are still some coreference errors in the generated stories. In order to encourage model to attend correct entities, we add a supervision on attention weight of entity mention tokens. We use Stanford's CoreNLP tool \footnote{https://stanfordnlp.github.io/CoreNLP/} to extract coreference annotation of stories.

Specifically, for a story $\mathbf{Y}$ we get $p$ coreference clusters and each cluster contains $q$ entity mentions. We assign each entity mention token $y^c_i$ in subsequence $\mathbf{Y}^c=\{y^c_1,y^c_i,y^c_{pq}\}$ a cluster label $\mathbf{C}=\{c_1,c_i,c_{pq}\}$. During training, for a entity mention token $y^c_i$, we get attention weights between current token and previous tokens $\{y^c\leq i-1\}$ in last self-attention layer of decoder, the sum of which is 1:
\begin{equation}
	\sum_{k=1}^{i-1}{\alpha_{ik}} = 1
\end{equation}
We design a coreference loss to  maximize attention weights of tokens in the same cluster as follows:
\begin{equation}
	\mathcal{L}_{\mathrm{coref}} = -\frac{1}{pq} \sum^{pq}_{i=1} \frac{1}{N_{i}}\sum_{k=1}^{i-1} \mathbbm{1}(c_k=c_i)  \log\alpha_{ik}
\end{equation}
where $N_{t}$ is the number of entity mentions in the same cluster $c_i$. Considering these two components, the loss function for the second stage model is as follows:
\begin{equation}
	\mathcal{L}=\mathcal{L}_{\mathrm{lm}}+ \lambda_1\mathcal{L}_{\mathrm{dis}} + \lambda_2\mathcal{L}_{\mathrm{coref}}
\end{equation}

% \begin{table}[!t]
	
% 	\centering
% % 	\resizebox{0.4\columnwidth}{!}{
% 	\begin{tabular}{llll}
% 		\toprule
% 		\textbf{Dataset}  & TRAIN   & VAL    & TEST   \\
% 		\midrule
% 		Number of stories & 272,541 & 15,619 & 15,136 \\
% 		Avg Len of prompt & 28.4    & 29.0   & 28.1   \\
% 		Avg Len of story  & 397.3   & 395.8  & 398.2  \\
% 		\bottomrule
% 	\end{tabular}
% % 	}
% 		\caption{Statistics of the dateset.}
% 		\label{statistics}
% \end{table}

\section{Experimental Setup}

%In this section, we describe our experiments on a real story dataset. We first introduce the dataset for this task. Then we compare our model with four baselines. We use automatic evaluation and human evaluation to evaluate our model and detail results analysis.

\subsection{Settings and Data Set}
% \subsection{Settings}
For two Transformer decoders, we apply the same model size as GPT2-117M \cite{radford2019language}. Thus we can analysis the effect of pre-training weight of GPT2. Specifically,  the dimension of word embedding and the dimension of hidden vectors are set to 768. The number of self attention block is set to 12 and 12 heads are used in self multi-head attention. We train the model using Adam \cite{kingma2014adam} with learning rate 0.0005. The dropout rate is set to 0.3 for regularization.  $\lambda_1$ and $\lambda_2$ are set to 0.1 and 0.3 according to the performance in valid set. Following \cite{fan2018hierarchical} we generate stories with random top $k$ sampling, where next words are sampling from the top $k=20$ candidates rather than the entire vocabulary distribution.

% \subsection{Data Set}
We use writing prompts dataset from \cite{fan2018hierarchical}, which is collected from Reddit's WRITINGPROMPTS forum\footnote{https://www.reddit.com/r/WritingPrompts/}. WRITINGPROMPTS is a community where online users inspire each other to write by submitting story prompts. Each prompt can have multiple story responses. The prompts have a large diversity of topic, length, and detail. There are 300k stories and the dataset is split into TRAIN, VAL and TEST (90\%/5\%/5\%). For our experiments, we limit the length of the stories to 500 words maximum. We use the GPT2's BPE vocabulary with size of 50,527 in our model. 
% Statistics of the dateset are given in Table \ref{statistics}.

\subsection{Evaluation Metrics}

\noindent\textbf{Automatic Evaluation.}
Many commonly used metrics based on n-gram overlap between the generated text and the human text, such as BLEU \cite{papineni2002bleu}, are not useful in story generation, which is also observed by previous work \cite{martin2018event,fan2018hierarchical}. Because we do not aim to generate a specific story; we want to generate viable and novel stories. 

In order to evaluate different aspect of stories we use four type metrics. We use \textbf{Perplexity} to evaluate the fluency of stories. Perplexity is commonly used to evaluate the quality of language models, and it reflects how fluently the model can produce the correct next word given the preceding words. What's more, in order to evaluate the diversity of stories we compute \textbf{Distinct-1/2} \cite{li2016persona}, which is the percentage of distinct n-grams in all generated stories and is widely used in conversation generation. 

In order to evaluate the discourse coherency of the stories, we reuse the fine-tuned BERT for evaluation. Specifically, we use BERT to tag discourse label for sentence pairs in generated stories in the same way as the tagging process of training set in Section \ref{sec:discourse}. We compute the percentage of sentence pairs with \textbf{Unknown} labels in generated stories. The less sentence pairs with unknown labels the model generates, the better the coherency of stories are.  In order to evaluate the coreference coherence, we compute the averaged \textbf{Coreference Chains} in each story. Specifically, we use Stanford's CoreNLP tool \footnote{https://stanfordnlp.github.io/CoreNLP/} to extract coreference chains of generated stories. 

\noindent\textbf{Human Evaluation.}
To further evaluate the quality of generated stories, we conduct pair-wise comparisons with two strong baseline models (FConvS2S and GPT2P). 
We evaluate the models from the following three perspectives: \textbf{Relevance} to indicate whether a story is relevant to the given prompt, \textbf{Grammaticality} to indicate whether a story is natural and
fluent, and \textbf{Logicality} to indicate whether a story is consistent and coherent in terms of causal dependencies in the context. Three aspects are independently evaluated.
We randomly sample 100 stories from the test set and obtain 300 stories from three models. For each pair of stories (one by our model and the other by a baseline, along with the prompt), three annotators are asked to give a preference (win, lose, or tie) in terms of three metrics respectively. Majority voting is used to make final decisions among the three annotators.

\subsection{Comparison Methods}

\noindent\textbf{Conv Seq2Seq with self-attention (ConvS2S).} We replicate the model proposed by \cite{fan2018hierarchical} using their source code, which applies a convolutional sequence-to-sequence model with gated self-attention to generate stories from prompts.

\noindent\textbf{Fusion of Conv Seq2Seq with self-attention (FConvS2S).} The model is also proposed by \cite{fan2018hierarchical}, which utilizes a fusion mechanism to integrate two \textbf{ConvS2S}.

\noindent\textbf{GPT2.} The model only contains a Transformer-based decoder and has the same model size as GPT2-117M \cite{radford2019language}. We train the model from scratch.

\noindent\textbf{GPT2 with Pre-training (GPT2P).} We first load pre-training weights of GPT2-117M and then fine tune the model on the used dataset.

% \noindent\textbf{TLM+Discourse.} This model enhances Transformer based language model with our proposed discourse relation classification component.

% \noindent\textbf{TLM+Coherence.} This model enhances Transformer based language model with our proposed coherence supervision component.

\noindent\textbf{Ours.} Our overall model contains two-stage generation, discourse relation classification and coreference supervision. In order to evaluate the upper bound of two-stage generation, we use different percentages of tokens of ground truth outlines as contexts to generate stories. Ours(0\%) means using own generated outlines as contexts in the second stage to generate stories. It is our final model. Ours(100\%) means all tokens of ground truth outlines are used as contexts. It is the upper bound  model.

\section{Results and Discussions}

\vspace{-2mm}
\subsection{Automatic Evaluation and Human Evaluation}

\begin{table*}[!htb]
	
	\centering
	\resizebox{2\columnwidth}{!}{
	\begin{tabular}{lccccc}
		\toprule
		\textbf{Method} & \textbf{Perplexity}$\downarrow$ & \textbf{Dis-1}(\%)$\uparrow$ & \textbf{Dis-2}(\%)$\uparrow$  & \textbf{Unknown}(\%)$\downarrow$ & \textbf{Coref Chains}$\uparrow$ \\
		\midrule
		ConvS2S   & 34.61   & 0.400              & 5.191           & 76.01          & 5.52                  \\
		FConvS2S  & 33.97   & 0.482              & 6.271           & 75.60          & 5.43                  \\
		GPT2      & 29.50   & 0.474              & 6.796           & 74.95          & 5.67                  \\
		GPT2P     & 25.64   & 0.493              & 7.333           & 73.61          & 5.61                  \\
        % Ours      & 10.32   & \textbf{0.530}     & \textbf{7.379}   & 74.58           & \textbf{5.98}       \\
        % Ours      & 10.32/31.42   & 1.509/0.530     & 15.266/7.379  & 74.97/74.58          & 5.79/5.98      \\
        Ours(0\% ground truth outline)      & 30.84   & 0.531     & 7.379    & 75.19           & 5.98       \\
       \midrule
        Ours(50\% ground truth outline)      & 19.21   & 1.311     & 13.253   & 75.15           & 5.97      \\
        Ours(100\% ground truth outline)      & 10.32   & 1.509    & 15.266   & 74.97           & 5.80       \\
		\bottomrule
	\end{tabular}
	}
	\caption{Automatic evaluation results on TEST set.}
	\label{overallresult}
% 	\vspace{-15mm}
\end{table*}

\begin{table*}[!htb]
\centering
	\resizebox{2\columnwidth}{!}{
\begin{tabular}{llllllllll}
\toprule
\multirow{2}{*}{\textbf{Method}} & \multicolumn{3}{l}{\textbf{Relevance}} & \multicolumn{3}{l}{\textbf{Grammaticality}} & \multicolumn{3}{l}{\textbf{Logicality}} \\
                        & \textbf{Win}(\%)  & \textbf{Tie}(\%) & \textbf{Lose}(\%) & \textbf{Win}(\%)   & \textbf{Tie}(\%)   & \textbf{Lose}(\%)   & \textbf{Win}(\%)  & \textbf{Tie}(\%)  & \textbf{Lose}(\%) \\
\midrule
Ours vs. FConvS2S       & \textbf{23}       & 66      & 11       & \textbf{28}        & 53        & 19         & \textbf{40}       & 33       & 27       \\
Ours vs. GPT2P          & \textbf{21}       & 60      & 19       & \textbf{17}        & 69        & 14         & \textbf{31}       & 47       & 22       \\
% Ours vs. TLM-T          & 27       & 53      & 20       & 15        & 72        & 13         & 28       & 51       & 21 \\
\bottomrule
\end{tabular}
}
\caption{Human evaluation results on TEST set.}
	\label{humanresult}
% 	\vspace{-9mm}
\end{table*}

\begin{table*}[!htb]
	
	\centering
	\resizebox{1.8\columnwidth}{!}{
	\begin{tabular}{llllll}
		\toprule
		\textbf{Method} & \textbf{Perplexity}$\downarrow$  & \textbf{Dis-1}(\%)$\uparrow$ & \textbf{Dis-2}(\%)$\uparrow$ & \textbf{Unknown}(\%)$\downarrow$ & \textbf{Coref Chains}$\uparrow$ \\
		\midrule
		\textbf{First stage}                                                                                         \\
		\midrule
		keyword         & 74.46   & 0.964              & 7.132      &/ &/        \\
		abstract        & 35.53   & 0.776              & 10.060     &/ &/         \\
		\midrule
		\textbf{Second stage}                                                                                        \\
		\midrule
		story with keyword         & 17.82   & 0.461              & 6.188     & 74.26  & 5.67          \\
		story with abstract        & 10.65   & 0.512              & 7.358     & 74.54  & 5.81         \\
		\bottomrule
	\end{tabular}
	}
	\caption{Comparison of different outlines.}
	\label{twostageresult}
% 	\vspace{-12mm}
\end{table*}

As shown in Table \ref{overallresult}, we compute four types metrics for these methods. We can see that GPT2 outperforms FConvS2S and ConvS2S in all metrics. This indicates that the self-attention based model is superior to the convolutional based model in story generation. Although FConvS2S and ConvS2S is enhanced with a self-attention mechanism, their ability to capture long-distance dependence is still weaker than GPT2. Compared to GPT2, GPT2P improves the perplexity and distinct significantly. GPT2P also generates least sentence pairs with \textit{unknown} discourse relation. This shows that pre-training weights contributions to generating more fluent, diverse and coherent stories.  Compared to these methods, our model (Ours(0\%)) achieves best diversity and coreference performance. This demonstrates the effectiveness of our overall model. The upper bound model (Ours(100\%)) achieves best perplexity score. This indicates that our model sacrifices part of fluency for the plot control. 
What's more, we can see that all two-stage models has a lower \textit{unknown} score compared with GPT2 and GPT2P. We claim that two-stage generation and discourse relation component may repel each other. Next, we conduct ablation experiment to evaluate each component of our method.

Table \ref{humanresult} reports human evaluation results.
Our method achieves best scores in three metrics. Specifically, our method mainly improve scores on Logicality. This shows
that our method can generate more coherent stories by utilizing discourse and coreference supervision. Our method performs similar to GPT2P in term of Relevance and Grammaticality. Because both two methods use Transformer as the decoder and our model dose not design a component to improve the relevance to the prompt.

% \vspace{-6mm}
\subsection{Outline Analysis}

\begin{figure}[!ht]
	\centering
	\includegraphics[width=7cm]{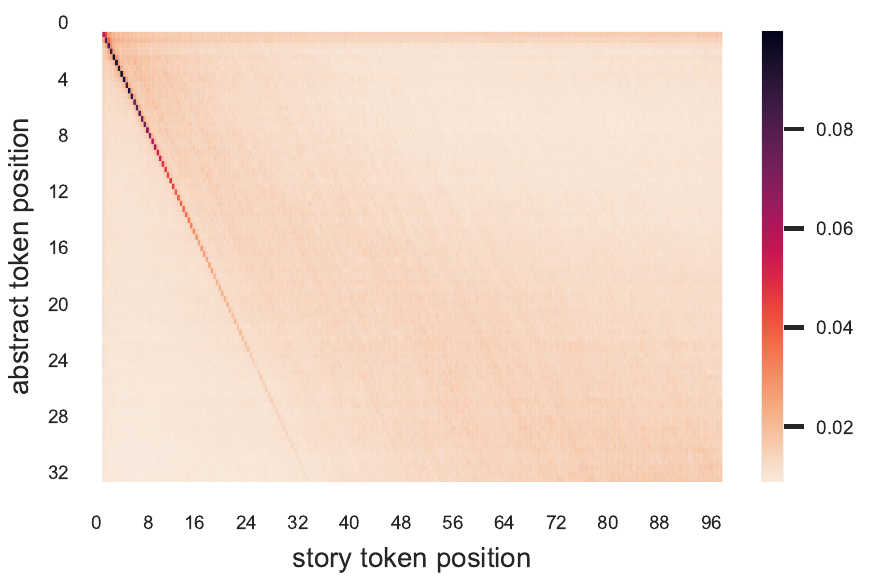}
	\caption{The attention weight distribution of story tokens in different positions.}
	\label{abstract}
% 	\vspace{-6mm}
\end{figure}
We compare the performance of keyword and abstract as outlines. As shown in Table \ref{twostageresult}, in first stage keyword is more difficult to generate than abstract, for that keyword gets a higher perplexity. From second stage, we can see that stories using abstract as outline get better scores in four metrics. This indicates that the abstract contributes to generating stories with better diversity and consistency. Therefore, we take abstract as outline in our model. In order to evaluate whether the stories are generated following the plot order of abstract, we plot story tokens' attention weight distributions on abstract tokens. The attention weight distributions are computed by averaging 2,000 generated stories. Because of the limited space, we only list tokens of the abstract and the story in the front positions. The result is shown in Figure \ref{abstract}. There are several lines with darker colors in the diagonal direction of the figure. This demonstrates that the story's focus follows the plot order of the abstract and our two-stage model can control the plots of the story well.

\subsection{Discourse Relation Classification}

\begin{table}[!htb]
	\centering
	\resizebox{1\columnwidth}{!}{
\begin{tabular}{l|lll}
	\toprule
	\textbf{TLM+Discourse} & \textbf{And}(\%)$\uparrow$ & \textbf{When}(\%)$\uparrow$ & \textbf{Unknown}(\%)$\downarrow$ \\
	\midrule
	0.1        & 11.43      & 2.90    & 72.94      \\
	0.3        & 11.38      & 2.80    & 73.60      \\
	0.5        & 10.91      & 2.72    & 73.78      \\
	\bottomrule
\end{tabular}
}
	\caption{The percentages of discourse relations with different $\lambda_1$.}
	\label{table:hyper}
% 	\vspace{-16mm}
\end{table}

\begin{table}[!htb]
	\centering
	\resizebox{1\columnwidth}{!}{
\begin{tabular}{l|lll}
	\toprule
	\textbf{Method} & \textbf{And}(\%)$\uparrow$ & \textbf{When}(\%)$\uparrow$ & \textbf{Unknown}(\%)$\downarrow$ \\
	\midrule
	ConvS2S        & 8.52  & 2.45 & 76.01 \\ 
	FConvS2S       & 8.67  & 2.41 & 75.59 \\ 
	GPT2           & 10.14 & 2.50 & 74.96 \\  
	GPT2P          & 10.96 & 2.60 & 73.61 \\  
	TLM+Discourse  & 11.43 & 2.90 & 72.94 \\
	\bottomrule
\end{tabular}
}
	\caption{The percentages of discourse relations of different methods.}
	\label{table:disresult}
% 	\vspace{-8mm}
\end{table}

% \begin{figure}[!ht]
% 	\centering
% 	\includegraphics[width=6cm]{all_discourse4.pdf}
% 	\caption{The percentages of discourse relations of different methods.}
% 	\label{discourse}
% \end{figure}

We evaluate the effect of discourse relation classification with \textbf{TLM+Discourse}, which enhances Transformer based language model with our proposed discourse relation classification component. We show the percentages of \textit{and}, \textit{when} and \textit{unknown} for that the percentages of other relations are small. We first try several different values for $\lambda_1$ in loss function and the result is shown in Table \ref{table:hyper}. When $\lambda_1$ is high, there is no gain in three discourse metrics. When $\lambda_1$ is 0.1, we get most gain in three discourse metrics. This indicates that large weight affects the main task and damages the performance of language model. Therefore, we use 0.1 in our final model. Then we compare \textbf{TLM+Discourse} with other baselines. The percentages of discourse relations are shown in Figure \ref{table:disresult}. We can see that GPT2 and GPT2P performs better than ConvS2S and FConvS2S. TLM+Discourse generates least sentence pairs with \textit{unknown} relation and achieves best score among all methods. This shows that discourse relation classification component is helpful to generate more sentence pairs with certain discourse relations and further improve the coherency of generated stories. 

% As shown in Table \ref{overallresult}, our model with only discourse component (TLM+Discourse) generates lease sentence pairs with \textit{unknown} discourse relation (72.94\%) compared to other baselines. This indicates our discourse relation classification model helps to generate stories with better discourse coherence. 
% \subsection{Coreference Supervision}
% As shown in Table \ref{overallresult}, our model with only coreference supervision generate stories with more coreference chains that GPT2P. This indicates that the extra coreference supervision helps the model to attend on entities more accurately and generate stories with better coreference consistency.

\subsection{Case Study}

\begin{table}[!htb]
	
	\centering
% 	\resizebox{2\columnwidth}{!}{
\begin{tabular}{p{0.95\columnwidth}}  
	\toprule
	\textbf{Prompt}: \small The zombie apocalypse is upon us but all humans are immune to the virus . All other mammals are susceptible . \\
	\midrule
	\small \textbf{GPT2P}: As I sat there I tried to remember how many days it had been since the last one died . $[$\textcolor{blue}{My wife and daughter were dead , I knew they would n't be alive any time soon .}$]_{1a}$ The food ran out quickly , \textcolor{red}{they} had to be taken away to stay in the house , but that was no reason to be angry if there were a few people left living there . After a month of food being gone a little girl sat next to me . She was the only other survivor who was n't bitten , as it was hard for us to eat \textcolor{red}{them} , we were all sickly . That would be another month to go and if the other animals were still alive in the food , they would n't be able to eat her . `` Hey little one . '' I looked up and saw her . $[$\textcolor{blue}{I had a big smile on my face at this point .}$]_{2a}$ $[$\textcolor{blue}{My wife was the only one who noticed , but her smile was all that was visible of me , and not of her .}$]_{1b}$ $[$\textcolor{blue}{I looked at her , and smiled sadly .}$]_{2b}$ She looked sad , but did n't care . She never cared for me .       \\
    \midrule
    \small \textbf{Our (generated abstract)}:  He was the last to see the end of this world . The only reason John had survived was by his own hand , and it took much more than he could bear with a few months to heal his wounds . It took a few months for him to be able to see the end and get his arm back from the dead , and he did n't care for a few more years without having to go back . \\
	\small \textbf{Our (generated story)}:  When he had been first laid to rest the first few days he did n't care , and he had thought it was just a small thing , but it started getting worse . He was the last to see the end of this world . The only reason John had survived was by his own hand , and it took much more than he could bear with a few months to heal his wounds . John had lived the rest of his life on Earth without anything else but the clothes on his back , and the rifle he kept at home when he was alone . The first few days were easy . A little bit of food , a bit of protection . He spent a few days walking around the city and drinking a beer , thinking he was safe , but the next few months were hell . It took a few months for him to be able to see the end and get his arm back from the dead , and he did n't care for a few more years without having to go back . It was better to go back , to be safe , so he would be safe for a while , and so he would n't get infected . 
       \\
	\bottomrule
\end{tabular}
\caption{Comparison of different methods.}
	\label{human}
% 	\vspace{-8mm}
% }
\end{table}

We analyze some generated stories to evaluate our overall model. An example is shown in Table \ref{human}. Stories generated by GPT2P have poor plot consistency and have some coreference errors, such as blue sentences and red words in Table \ref{human}. Compared with GPT2P, our model can effectively control the plot consistency of the story through the abstract. Therefore, stories generated by our model have better plot consistency. In addition, our model has less coreference errors than GPT2P and generates stories with better coreference consistency. What's more, the coherency between sentences is also better than GPT2P.

\section{Conclusion}

In this paper, we propose a two-stage generation model to improve consistency and coherency of generated stories. The first stage is to build the story outline, and the second stage is to expand the outline into a complete story. What's more, we design a supplementary task of discourse relation classification to improve the discourse representation ability of the model. In addition, we enhance model with coreference supervision to improve coreference consistency in generated stories. Experimental results on a story dataset show that our method is superior to baseline methods.

% There are some directions of improvement. We will introduce additional background knowledge to improve consistency of outline, such as CommonConcept and WordNet. We will use reinforcement learning to jointly optimize outline generation and story generation.

% include your own bib file like this:
\bibliographystyle{acl_natbib}
\bibliography{coling2020}

\begin{thebibliography}{32}
\expandafter\ifx\csname natexlab\endcsname\relax\def\natexlab#1{#1}\fi

\bibitem[{Bahdanau et~al.(2014)Bahdanau, Cho, and Bengio}]{bahdanau2014neural}
Dzmitry Bahdanau, Kyunghyun Cho, and Yoshua Bengio. 2014.
\newblock Neural machine translation by jointly learning to align and
  translate.
\newblock \emph{arXiv preprint arXiv:1409.0473}.

\bibitem[{Bai and Zhao(2018)}]{bai-zhao-2018-deep}
Hongxiao Bai and Hai Zhao. 2018.
\newblock Deep enhanced representation for implicit discourse relation
  recognition.
\newblock In \emph{Proceedings of the 27th International Conference on
  Computational Linguistics}, pages 571--583, Santa Fe, New Mexico, USA.
  Association for Computational Linguistics.

\bibitem[{Barrios et~al.(2016)Barrios, L{\'o}pez, Argerich, and
  Wachenchauzer}]{barrios2016variations}
Federico Barrios, Federico L{\'o}pez, Luis Argerich, and Rosa Wachenchauzer.
  2016.
\newblock Variations of the similarity function of textrank for automated
  summarization.
\newblock \emph{arXiv preprint arXiv:1602.03606}.

\bibitem[{Brown et~al.(2020)Brown, Mann, Ryder, Subbiah, Kaplan, Dhariwal,
  Neelakantan, Shyam, Sastry, Askell, Agarwal, Herbert-Voss, Krueger, Henighan,
  Child, Ramesh, Ziegler, Wu, Winter, Hesse, Chen, Sigler, Litwin, Gray, Chess,
  Clark, Berner, McCandlish, Radford, Sutskever, and
  Amodei}]{brown2020language}
Tom~B. Brown, Benjamin Mann, Nick Ryder, Melanie Subbiah, Jared Kaplan,
  Prafulla Dhariwal, Arvind Neelakantan, Pranav Shyam, Girish Sastry, Amanda
  Askell, Sandhini Agarwal, Ariel Herbert-Voss, Gretchen Krueger, Tom Henighan,
  Rewon Child, Aditya Ramesh, Daniel~M. Ziegler, Jeffrey Wu, Clemens Winter,
  Christopher Hesse, Mark Chen, Eric Sigler, Mateusz Litwin, Scott Gray,
  Benjamin Chess, Jack Clark, Christopher Berner, Sam McCandlish, Alec Radford,
  Ilya Sutskever, and Dario Amodei. 2020.
\newblock \href {http://arxiv.org/abs/2005.14165} {Language models are few-shot
  learners}.

\bibitem[{Chen et~al.(2019)Chen, Liu, Luan, Zhang, Liu, and
  Sun}]{chen2019learning}
Gang Chen, Yang Liu, Huanbo Luan, Meng Zhang, Qun Liu, and Maosong Sun. 2019.
\newblock Learning to predict explainable plots for neural story generation.
\newblock \emph{arXiv preprint arXiv:1912.02395}.

\bibitem[{Chen et~al.(2016)Chen, Zhang, Liu, Qiu, and
  Huang}]{chen-etal-2016-implicit}
Jifan Chen, Qi~Zhang, Pengfei Liu, Xipeng Qiu, and Xuanjing Huang. 2016.
\newblock Implicit discourse relation detection via a deep architecture with
  gated relevance network.
\newblock In \emph{Proceedings of the 54th Annual Meeting of the Association
  for Computational Linguistics (Volume 1: Long Papers)}, pages 1726--1735,
  Berlin, Germany. Association for Computational Linguistics.

\bibitem[{Clark et~al.(2018)Clark, Ji, and Smith}]{clark2018neural}
Elizabeth Clark, Yangfeng Ji, and Noah~A Smith. 2018.
\newblock Neural text generation in stories using entity representations as
  context.
\newblock In \emph{Proceedings of the 2018 Conference of the North American
  Chapter of the Association for Computational Linguistics: Human Language
  Technologies, Volume 1 (Long Papers)}, pages 2250--2260.

\bibitem[{Devlin et~al.(2018)Devlin, Chang, Lee, and
  Toutanova}]{devlin2018bert}
Jacob Devlin, Ming-Wei Chang, Kenton Lee, and Kristina Toutanova. 2018.
\newblock Bert: Pre-training of deep bidirectional transformers for language
  understanding.
\newblock \emph{arXiv preprint arXiv:1810.04805}.

\bibitem[{Fan et~al.(2018)Fan, Lewis, and Dauphin}]{fan2018hierarchical}
Angela Fan, Mike Lewis, and Yann Dauphin. 2018.
\newblock Hierarchical neural story generation.
\newblock In \emph{Proceedings of the 56th Annual Meeting of the Association
  for Computational Linguistics (Volume 1: Long Papers)}, pages 889--898.

\bibitem[{Fan et~al.(2019)Fan, Lewis, and Dauphin}]{fan-etal-2019-strategies}
Angela Fan, Mike Lewis, and Yann Dauphin. 2019.
\newblock Strategies for structuring story generation.
\newblock In \emph{Proceedings of the 57th Annual Meeting of the Association
  for Computational Linguistics}, pages 2650--2660, Florence, Italy.
  Association for Computational Linguistics.

\bibitem[{Guan et~al.(2020)Guan, Huang, Zhao, Zhu, and
  Huang}]{guan2020knowledge}
Jian Guan, Fei Huang, Zhihao Zhao, Xiaoyan Zhu, and Minlie Huang. 2020.
\newblock A knowledge-enhanced pretraining model for commonsense story
  generation.
\newblock \emph{Transactions of the Association for Computational Linguistics},
  8:93--108.

\bibitem[{Jain et~al.(2017)Jain, Agrawal, Mishra, Sukhwani, Laha, and
  Sankaranarayanan}]{jain2017story}
Parag Jain, Priyanka Agrawal, Abhijit Mishra, Mohak Sukhwani, Anirban Laha, and
  Karthik Sankaranarayanan. 2017.
\newblock Story generation from sequence of independent short descriptions.
\newblock \emph{arXiv preprint arXiv:1707.05501}.

\bibitem[{Kingma and Ba(2014)}]{kingma2014adam}
Diederik~P Kingma and Jimmy Ba. 2014.
\newblock Adam: A method for stochastic optimization.
\newblock \emph{arXiv preprint arXiv:1412.6980}.

\bibitem[{Lan et~al.(2017)Lan, Wang, Wu, Niu, and Wang}]{lan-etal-2017-multi}
Man Lan, Jianxiang Wang, Yuanbin Wu, Zheng-Yu Niu, and Haifeng Wang. 2017.
\newblock Multi-task attention-based neural networks for implicit discourse
  relationship representation and identification.
\newblock In \emph{Proceedings of the 2017 Conference on Empirical Methods in
  Natural Language Processing}, pages 1299--1308, Copenhagen, Denmark.
  Association for Computational Linguistics.

\bibitem[{Lebowitz(1987)}]{lebowitz1987planning}
Michael Lebowitz. 1987.
\newblock Planning stories.
\newblock In \emph{Proceedings of the 9th annual conference of the cognitive
  science society}, pages 234--242.

\bibitem[{Li et~al.(2013)Li, Lee-Urban, Johnston, and Riedl}]{li2013story}
Boyang Li, Stephen Lee-Urban, George Johnston, and Mark Riedl. 2013.
\newblock Story generation with crowdsourced plot graphs.
\newblock In \emph{Twenty-Seventh AAAI Conference on Artificial Intelligence}.

\bibitem[{Li et~al.(2016)Li, Galley, Brockett, Spithourakis, Gao, and
  Dolan}]{li2016persona}
Jiwei Li, Michel Galley, Chris Brockett, Georgios Spithourakis, Jianfeng Gao,
  and Bill Dolan. 2016.
\newblock A persona-based neural conversation model.
\newblock In \emph{Proceedings of the 54th Annual Meeting of the Association
  for Computational Linguistics (Volume 1: Long Papers)}, pages 994--1003.
  Association for Computational Linguistics.

\bibitem[{Martin et~al.(2018)Martin, Ammanabrolu, Wang, Hancock, Singh,
  Harrison, and Riedl}]{martin2018event}
Lara~J Martin, Prithviraj Ammanabrolu, Xinyu Wang, William Hancock, Shruti
  Singh, Brent Harrison, and Mark~O Riedl. 2018.
\newblock Event representations for automated story generation with deep neural
  nets.
\newblock In \emph{Thirty-Second AAAI Conference on Artificial Intelligence}.

\bibitem[{Mostafazadeh et~al.(2016)Mostafazadeh, Chambers, He, Parikh, Batra,
  Vanderwende, Kohli, and Allen}]{mostafazadeh-etal-2016-corpus}
Nasrin Mostafazadeh, Nathanael Chambers, Xiaodong He, Devi Parikh, Dhruv Batra,
  Lucy Vanderwende, Pushmeet Kohli, and James Allen. 2016.
\newblock A corpus and cloze evaluation for deeper understanding of commonsense
  stories.
\newblock In \emph{Proceedings of the 2016 Conference of the North {A}merican
  Chapter of the Association for Computational Linguistics: Human Language
  Technologies}, pages 839--849, San Diego, California. Association for
  Computational Linguistics.

\bibitem[{Nie et~al.(2019)Nie, Bennett, and Goodman}]{nie-etal-2019-dissent}
Allen Nie, Erin Bennett, and Noah Goodman. 2019.
\newblock {D}is{S}ent: Learning sentence representations from explicit
  discourse relations.
\newblock In \emph{Proceedings of the 57th Annual Meeting of the Association
  for Computational Linguistics}, pages 4497--4510, Florence, Italy.
  Association for Computational Linguistics.

\bibitem[{Papineni et~al.(2002)Papineni, Roukos, Ward, and
  Zhu}]{papineni2002bleu}
Kishore Papineni, Salim Roukos, Todd Ward, and Wei-Jing Zhu. 2002.
\newblock Bleu: a method for automatic evaluation of machine translation.
\newblock In \emph{Proceedings of the 40th annual meeting on association for
  computational linguistics}, pages 311--318. Association for Computational
  Linguistics.

\bibitem[{P{\'E}rez and Sharples(2001)}]{perez2001mexica}
Rafael P{\'E}rez~{\'Y} P{\'E}rez and Mike Sharples. 2001.
\newblock Mexica: A computer model of a cognitive account of creative writing.
\newblock \emph{Journal of Experimental \& Theoretical Artificial
  Intelligence}, 13(2):119--139.

\bibitem[{Porteous and Cavazza(2009)}]{porteous2009controlling}
Julie Porteous and Marc Cavazza. 2009.
\newblock Controlling narrative generation with planning trajectories: the role
  of constraints.
\newblock In \emph{Joint International Conference on Interactive Digital
  Storytelling}, pages 234--245. Springer.

\bibitem[{Prasad et~al.(2008)Prasad, Dinesh, Lee, Miltsakaki, Robaldo, Joshi,
  and Webber}]{prasad-etal-2008-penn}
Rashmi Prasad, Nikhil Dinesh, Alan Lee, Eleni Miltsakaki, Livio Robaldo,
  Aravind Joshi, and Bonnie Webber. 2008.
\newblock The {P}enn discourse {T}ree{B}ank 2.0.
\newblock In \emph{Proceedings of the Sixth International Conference on
  Language Resources and Evaluation ({LREC}'08)}, Marrakech, Morocco. European
  Language Resources Association (ELRA).

\bibitem[{Radford et~al.(2019)Radford, Wu, Child, Luan, Amodei, and
  Sutskever}]{radford2019language}
Alec Radford, Jeffrey Wu, Rewon Child, David Luan, Dario Amodei, and Ilya
  Sutskever. 2019.
\newblock Language models are unsupervised multitask learners.
\newblock \emph{OpenAI Blog}, 1(8).

\bibitem[{Riedl and Young(2010)}]{riedl2010narrative}
Mark~O Riedl and Robert~Michael Young. 2010.
\newblock Narrative planning: Balancing plot and character.
\newblock \emph{Journal of Artificial Intelligence Research}, 39:217--268.

\bibitem[{Rose et~al.(2010)Rose, Engel, Cramer, and Cowley}]{rose2010automatic}
Stuart Rose, Dave Engel, Nick Cramer, and Wendy Cowley. 2010.
\newblock Automatic keyword extraction from individual documents.
\newblock \emph{Text mining: applications and theory}, 1:1--20.

\bibitem[{See et~al.(2019)See, Pappu, Saxena, Yerukola, and
  Manning}]{see-etal-2019-massively}
Abigail See, Aneesh Pappu, Rohun Saxena, Akhila Yerukola, and Christopher~D.
  Manning. 2019.
\newblock Do massively pretrained language models make better storytellers?
\newblock In \emph{Proceedings of the 23rd Conference on Computational Natural
  Language Learning (CoNLL)}, pages 843--861, Hong Kong, China. Association for
  Computational Linguistics.

\bibitem[{Sutskever et~al.(2014)Sutskever, Vinyals, and
  Le}]{sutskever2014sequence}
Ilya Sutskever, Oriol Vinyals, and Quoc~V Le. 2014.
\newblock Sequence to sequence learning with neural networks.
\newblock In \emph{Advances in neural information processing systems}, pages
  3104--3112.

\bibitem[{Xu et~al.(2018)Xu, Ren, Zhang, Zeng, Cai, and Sun}]{xu2018skeleton}
Jingjing Xu, Xuancheng Ren, Yi~Zhang, Qi~Zeng, Xiaoyan Cai, and Xu~Sun. 2018.
\newblock A skeleton-based model for promoting coherence among sentences in
  narrative story generation.
\newblock In \emph{Proceedings of the 2018 Conference on Empirical Methods in
  Natural Language Processing}, pages 4306--4315.

\bibitem[{Yang et~al.(2019)Yang, Dai, Yang, Carbonell, Salakhutdinov, and
  Le}]{yang2019xlnet}
Zhilin Yang, Zihang Dai, Yiming Yang, Jaime Carbonell, Ruslan Salakhutdinov,
  and Quoc~V Le. 2019.
\newblock Xlnet: Generalized autoregressive pretraining for language
  understanding.
\newblock \emph{arXiv preprint arXiv:1906.08237}.

\bibitem[{Yao et~al.(2019)Yao, Peng, Weischedel, Knight, Zhao, and
  Yan}]{yao2019plan}
Lili Yao, Nanyun Peng, Ralph Weischedel, Kevin Knight, Dongyan Zhao, and Rui
  Yan. 2019.
\newblock Plan-and-write: Towards better automatic storytelling.
\newblock In \emph{Proceedings of the AAAI Conference on Artificial
  Intelligence}, volume~33, pages 7378--7385.

\end{thebibliography}

\end{document}